\documentclass{article}

\usepackage{arxiv}

\usepackage[utf8]{inputenc} 
\usepackage[T1]{fontenc}    
\usepackage{hyperref}       
\usepackage{url}            
\usepackage{booktabs}       
\usepackage{amsfonts}       
\usepackage{nicefrac}       
\usepackage{microtype}      
\usepackage{lipsum}
\usepackage{amssymb}
\usepackage{amsmath}
\usepackage{algorithm}
\usepackage[noend]{algpseudocode}
\usepackage{graphicx,wrapfig}
\usepackage{caption}
\usepackage{cases}
\usepackage{subcaption}

\usepackage{xcolor}
\usepackage{dblfloatfix}    
\usepackage{tabularx,ragged2e,booktabs,caption}
\newcolumntype{C}[1]{>{\Centering}m{#1}}

\usepackage{amsthm}

\newtheorem{definition}{Definition}

\usepackage[flushleft]{threeparttable}

\title{Learn to Exceed: Stereo Inverse Reinforcement Learning with Concurrent Policy Optimization}

\author{
  Feng Tao \\
  Department of Electrical and Computer Engineering\\
  University of Texas\\
  San Antonio, TX 78249 \\
  \texttt{feng.tao@my.utsa.edu} \\
   \And
 Yongcan Cao \\
  Department of Electrical and Computer Engineering\\
  University of Texas\\
  San Antonio, TX 78249 \\
  \texttt{yongcan.cao@utsa.edu} 
}

\begin{document}
\maketitle

\begin{abstract}
In this paper, we study the problem of obtaining a control policy that can mimic and then outperform expert demonstrations in Markov decision processes where the reward function is unknown to the learning agent. One main relevant approach is the inverse reinforcement learning (IRL), which mainly focuses on inferring a reward function from expert demonstrations. The obtained control policy by IRL and the associated algorithms, however, can hardly outperform expert demonstrations. To overcome this limitation, we propose a novel method that enables the learning agent to outperform the demonstrator via a new concurrent reward and action policy learning approach. In particular, we first propose a new stereo utility definition that aims to address the bias in the interpretation of expert demonstrations. We then propose a loss function for the learning agent to learn reward and action policies concurrently such that the learning agent can outperform expert demonstrations. The performance of the proposed method is first demonstrated in OpenAI environments. Further efforts are conducted to experimentally validate the proposed method via an indoor drone flight scenario.
\end{abstract}

\keywords{Reinforcement learning; Inverse reinforcement learning; Robotics; Imitation to Exceed}

\section{Introduction}
Reinforcement learning (RL) has shown its advantages in yielding human-level or better-than-human-level performance in, \textit{e.g.}, Go and Atari games~\cite{silver2016mastering,mnih2015human}.
The basic idea of reinforcement learning is to learn control policies that optimize certain metrics. In many existing reinforcement learning algorithms, \textcolor{black}{such as DQN~\cite{mnih2015human}, REINFORCE~\cite{williams1992simple}, and proximal policy optimization (PPO)~\cite{schulman2017proximal}}, the reward function is used as the metric to evaluate the performance of a control policy. In particular, the deep Q-network (DQN) \textcolor{black}{leverages the immediate environment reward as part of its training label in the neural network updating process, where the neural network is used to approximate the Q values for each state. When the approximated Q values converge, an optimal action can then be derived.} The REINFORCE algorithms focus on approximating the policy directly by a neural network and adjusting the parameters directly based on the gradient of the cumulative reward, \textcolor{black}{\textit{i.e.}, summation of the discounted environment rewards along one trial}. The typical reinforcement learning algorithm used in the OpenAI Gym environment, namely, PPO, combines Q-learning with policy gradient methods in its policy search process based on the pre-defined environment rewards. Hence, the availability of reward functions plays a critical role in the design of reinforcement learning algorithms. 

In practice, however, the reward function itself may not often be available. Even in the synthetic cases when the reward function is pre-provided, research has proved that applying techniques like reward shaping~\cite{ng1999policy} can be beneficial to the policy search process. For example, \cite{kumar2018bipedal} observed that a proper shaped reward function can yield a faster walking agent than that trained with the original reward function, indicating that the original reward function may not be the optimal one. Moreover, the given reward function may suffer from corruption \cite{everitt2017reinforcement}. The noisy reward problem will consequently mislead the learning agent to some degree \cite{huang2019deceptive}. To overcome these challenges, it is important to learn task-specific reward functions for the design of optimal control policies that best fulfill the task objectives.

One important and popular approach for the learning/approximation of the reward function is the inverse reinforcement learning (IRL) \cite{ng2000algorithms, ziebart2008maximum, finn2016guided}, which seeks to learn the reward function from expert demonstrations. Despite recent significant advancements, the existing IRL methods still suffer from some limitations. First, experts are assumed to be available that can provide high-quality demonstrations, which is often difficult and time-consuming. Second, \textcolor{black}{the existing} IRL falls into the subcategory of imitation learning \cite{schaal1999imitation}, \textit{i.e.}, the learner's goal is to take actions via imitating the expert \cite{ho2016generative}. Hence, the learner could hardly outperform the expert. \textcolor{black}{To the best of our knowledge, there is one paper~\cite{syed2008game} that seeks to produce a policy that exceeds the performance of the expert from a game-theoretic view, assuming that the reward function is linear with respect to some given features. Such an assumption can also be found in the original IRL methods~\cite{zhifei2012review, ng2000algorithms, ziebart2008maximum, abbeel2004apprenticeship}. Although this assumption indeed simplifies the reward learning problem, it is impractical for real applications.} Some recent papers have focused on addressing nonlinear reward functions in order to overcome the limitations of linear model~\cite{levine2011nonlinear, wulfmeier2015maximum}. One limitation of these methods is that the proposed methods follow the common template for IRL \cite{arora2018survey}, which implies that the policy is optimized in the inner loop of reward function learning process. Because the policy optimization procedure can become very complex in high-dimensional systems, it is challenging and time-consuming to find an optimal policy based on the current reward function in the inner loop of reward function learning process. In \cite{finn2016guided}, a sample-based approach was formulated to approximate the partition function based on the maximum entropy IRL framework \cite{ziebart2008maximum}, which avoids the common IRL template and updates the reward function in the inner loop of policy search. However, the proposed approach in \cite{finn2016guided} can hardly provide better performance than expert demonstrations. 


To address these limitations, we propose a new reward and policy concurrent learning approach to recover the reward function and derive control policies that can mimic and then outperform expert demonstrations. In particular, we first introduce a stereo utility that calculates the expectation of the utility value for one trajectory/demonstration with respect to different discount factors. Because the discount factor used in the expert's decision making under Markov decision processes is unknown, averaging numerous discount factors can mitigate the bias on interpreting demonstrations. Furthermore, using a set of discount factors can provide a more comprehensive decision making process when future rewards are discounted heterogeneously. We then propose a new loss function aiming at enabling the learning agent to exceed the demonstrator. The loss function includes both the policy learning process and the reward function approximation process, hence yielding a concurrent learning structure. Finally, we demonstrate the effectiveness of the proposed algorithm in both synthetic and real-world environments.

\section{Preliminaries and Problem Statement}\label{sec:2}

\subsection{Reinforcement Learning Review}
A standard Markov decision process (MDP) can be represented as a tuple $\mathcal{M} := \langle \mathcal{S}, \mathcal{A}, \mathcal{T}, \mathcal{R}, \gamma \rangle$, where $\mathcal{S}$ denotes the state space, $\mathcal{A}$ denotes the action space, $ \mathcal{T}$ denotes the transition model, $\mathcal{R}$ denotes the reward function, and $\gamma \in [0,1]$ denotes the discount factor that is used to compute a weighted accumulation of past rewards for one trajectory in the form of $\{ s_0, a_0, r_1, ..., s_j, a_j, r_{j+1}, ...\}$, where $j\in \mathbb{N}$, $s_j \in \mathcal{S}$, and $a_j \in \mathcal{A}$. The selection of an action $a_j$ is determined by the action policy represented as $\pi_\theta(a_j|s_j): \mathcal{S} \rightarrow p(\mathcal{A}|\mathcal{S}, \theta)$, where $\theta$ is the parameter of the action policy generated, \textit{e.g.}, by a neural network. Note that $p(\cdot)$ can be either stochastic or deterministic. $r_j$ is an immediate reward for state $s_{j-1}$ after taking the action $a_{j-1}$, which is derived from the reward function $\mathcal{R}$. The standard reinforcement learning framework~\cite{sutton2018reinforcement} refers to the case when a learning agent interacts with such an MDP $\mathcal{M}$ in order to generate an optimal action policy such that the (discounted) accumulated reward is maximized. Since the reward function $\mathcal{R}$ in $\mathcal{M}$ is typically assumed to be known, the (discounted) cumulative reward, \textit{i.e.}, the (discounted) summation of the immediate rewards that one trajectory collects, is widely used in the existing RL approaches as the policy metric. When $\mathcal{R}$ is unknown or hard to be predetermined, a learned reward function is required in order to employ the existing RL approaches. One typical approach to learn/approximate the reward function is proposed via solving the inverse reinforcement learning (IRL) problem~\cite{russell1998learning, ng2000algorithms}. 


IRL studies the problem of learning an optimal reward function from expert demonstrations, which can be denoted as an MDP$\backslash{\mathcal{R}}$ problem~\cite{abbeel2004apprenticeship}. The expert is assumed to be attempting to optimize the cumulative reward when the reward function is assumed to be parameterized by given features. Let the parameters of the reward function be denoted as $\phi$ and the features be denoted as $x$. The reward function can be represented as $\mathcal{R}=f(x,\phi)$. The basic idea of IRL is to adjust $\phi$ so that the learned reward function can explain expert demonstrations well. In other words, the expert demonstrations should yield higher cumulative rewards than other randomly generated trajectories.

\subsection{Problem Statement}
Here we consider a new RL framework, which is defined as the interaction of a learning agent with an MDP whose reward function is \textit{unknown}. We denote it as a RL-MDP$\backslash{\mathcal{R}}$ problem. The new RL-MDP$\backslash{\mathcal{R}}$ is different from MDP$\backslash{\mathcal{R}}$ by including the simultaneous learning of action policies that aim to outperform the expert.


In this paper, we adopt the reward function specification~\cite{arora2018survey} as ${\mathcal{R}}: \mathcal{S}\rightarrow \mathbb{R}$, which provides a map from the state to the associated immediate reward~\cite{wulfmeier2015maximum}. To approximate the unknown relationship between states and rewards, a neural network architecture $g: \mathbb{R}^{n} \rightarrow \mathbb{R}$ can be used to take raw state features $s_{j} \in \mathcal{S}$, which could be high dimensional, as the input and return an immediate reward value. The approximated reward function $\hat{\mathcal{R}}$ is then represented as
\begin{equation}\label{eq:1}
    \hat{\mathcal{R}} := g(x| \phi),
\end{equation}
where $x \in \mathcal{S}$ and $\phi$ is the parameter associated with the neural network. We denote one sampled trajectory with an initial state $s_0$ in the RL-MDP$\backslash\mathcal{R}$ setting as $\tau^{\theta^+}_{s_0}$. The trajectory $\tau^{\theta{^+}}_{s_0}$ is sampled from the updated learner's policy $\pi_{\theta^+}$ (can be either stochastic or deterministic) governed by parameters $\theta^+$, such that
$\tau^{\theta^+}_{s_0} = \{s_0, \pi_{\theta^+}(a_0|s_0), s_1, ...,  \pi_{\theta^+}(a_{T-1}|s_{T-1}), s_{T}\},$
where $T$ represents the length of the sampled trajectory. Similarly, we denote the expert demonstration with the same initial state $s_0$ as 
$\tau^{\theta^*}_{s_0} = \{s_0, (a_0), s_1, (a_1) ..., s_{T^*-1}, (a_{T^*-1}), s_{T^*}\},$
where $T^*$ represents the length of the episode associated with the expert demonstration and the parentheses indicate that the action information is not required to be available (since $\hat{\mathcal{R}}$ defined in (\ref{eq:1}) only considers the raw state information as the features). If the initial state and the policy are not specified, the demonstration/trajectory notation can be simplified as $\tau^{\theta}$. 

To calculate the discounted cumulative reward for a given demonstration or trajectory, a standard approach~\cite{sutton2018reinforcement} is used in this paper. In particular, the discounted cumulative reward of one demonstration/trajectory $\tau^{\theta}$ is given by
$G_{\phi}(\tau^{\theta}, \gamma) = \sum_{j=1}^{T} \gamma^{j-1} g(s_{j}| \phi),$
where $\phi$ is the parameter of the reward function, $\gamma$ is the discount factor in the MDP, $s_j \in \tau^{\theta}$, and the neural network output value $g(s_{j}| \phi)$ is an estimation of the corresponding immediate reward ${r}_{j}$.

Note that the ground truth reward function $\mathcal{R}$ is unavailable for comparison. Our goal is to obtain learned reward functions $\hat{\mathcal{R}}$ such that (1) the policy learning process converges to a final policy that yields stable performance, and (2) $\hat{\mathcal{R}}$ can quantitatively distinguish trajectories, \textit{i.e.}, good trajectories yield larger discounted cumulative rewards than the bad ones.

\section{Reward and Policy Concurrent Learning} \label{sec:3}
In this section, we focus on proposing a new algorithm, named {reward and policy concurrent learning} (RPCL). We begin with a new cumulative reward calculation technique for the expert's demonstrations, named {stereo utility}. Then we construct a new loss function for the RL-MDP$\backslash\mathcal{R}$ problem. In particular, the new loss function can enable (1) the learning agent to exceed the demonstrator and (2) the concurrent update of the reward function and the action policy. We also provide the pseudocode of the proposed algorithm. 


\subsection{Stereo Utility}
The existing research on IRL usually chooses one single discount factor $\gamma$ for calculating the discounted cumulative rewards with respective to the expert's demonstrations. For example, \cite{ziebart2008maximum,shiarlis2016inverse} define $\gamma = 1$ for the finite length trajectories, with an inherent assumption that the expert takes the rewards accumulated in one trajectory equally. However, the expert's principle on viewing future rewards, \textit{i.e.}, $\gamma$, may be unavailable for the learner. Applying one single $\gamma$ that is different from the real discount factor, which can even be dynamic, that the expert uses will create a bias in quantifying trajectories. To mitigate this bias caused by the uncertainty of $\gamma$, we here define a new utility function to quantify the trajectory values by averaging the discounted cumulative rewards with different $\gamma$. We name the new measurement {stereo utility}, whose definition is given below.     
\begin{definition}\label{de:1}
For a set $\Gamma$ that contains different possible discount factors $\gamma$, the stereo utility of one trajectory $\tau^{\theta}$ is defined as 
\begin{equation}\label{eq:3}
    U_{\phi}(\tau^{\theta}) = \sum_{\gamma \in \Gamma} \frac{G_{\phi}(\tau^{\theta}, \gamma)}{|\Gamma|},
\end{equation} 
where $|\Gamma|$ is the cardinality of $\Gamma$ and $G_{\phi}(\tau^{\theta}, \gamma)$ is the discounted cumulative reward of trajectory $\tau^{\theta}$ under the discount factor $\gamma$. 
\end{definition}

The {stereo utility} can be interpreted as an expectation of the discounted cumulative reward with respect to the discount factor set $\Gamma$ (with a uniform distribution). If an appropriate $\Gamma$ is chosen, $U_{\phi}(\tau^{\theta^*}_{s_0})$ is expected to be a more appropriate function to explain the demonstration $\tau^{\theta^*}_{s_0}$ than the normal discounted cumulative reward $G_{\phi}(\tau^{\theta^*}_{s_0}, \gamma)$ when a fixed discount factor is selected. 





\subsection{Loss Function for RL-MDP$\backslash\mathcal{R}$}
The maximum margin optimization~\cite{abbeel2004apprenticeship,ratliff2006maximum}, which aims at learning a reward function \textcolor{black}{that makes the demonstrations work better than alternative policies by a margin}, is one of the foundational methods for IRL~\cite{arora2018survey}. We here adopt a similar formulation structure to tackle the RL-MDP$\backslash\mathcal{R}$ problem. However, the basic idea of maximum margin formulations is that the expert should always perform better than the learning agent, which contradicts our goal of training a superior learning agent. Hence, we propose a new structure that is different from the standard maximum margin formulations. The new formulation is named {unidirectional minimum margin formulation}, whose objective is to enable the learning agent to exceed the expert, if feasible. In particular, we unidirectionally minimize the {stereo utility} difference between the learning agent's current trajectory and the expert's demonstration with the same initial state $s_0$. In particular, the {unidirectional minimum margin formulation} is denoted as 
$
   \min_{\phi} {\Big (U_{\phi}(\tau^{\theta^+}_{s_0}) - U_{\phi}(\tau^{\theta^*}_{s_0})\Big)}, 
$
where $\phi$ is the parameter of the reward function, $\tau^{\theta^+}_{s_0}$ is the learning agent's trajectory sampled from an updated policy $\pi_{\theta^+}$, and $\tau^{\theta^*}_{s_0}$ denotes the expert's demonstration with \textcolor{black}{an initial state $s_0$, which does not need to be the same initial state as the learning agent's trajectory}. 

In the above formulation, we only minimize the {stereo utility} difference between the learning agent's current trajectory and the expert's demonstration unidirectionally so that the learning agent's performance is not fully bounded by the expert's demonstrations. In other words, the learning agent does not need to perfectly track the expert. However, the {stereo utility} difference may diverge when solving the optimization problem. In order to let the learning agent achieve better performance while avoiding the divergence, we further propose to revise it as
\begin{equation}\label{eq:5}
   \min_{\phi} \Big(\lambda {\big (U_{\phi}(\tau^{\theta^+}_{s_0}) - U_{\phi}(\tau^{\theta^*}_{s_0})\big)} - {G_{\phi}(\tau^{\theta}, \gamma)}\Big), 
\end{equation}
where $0 \leq \lambda < 1$ is the weight that determines the importance of minimizing the margin $U_{\phi}(\tau^{\theta^+}_{s_0}) - U_{\phi}(\tau^{\theta^*}_{s_0})$ and $G_{\phi}(\tau^{\theta}, \gamma)$ is the discounted cumulative reward of trajectory $\tau^{\theta}$ calculated with a predefined $\gamma$ in the policy optimization process. Note that $G_{\phi}(\tau^{\theta}, \gamma)$ can also be replaced by the stereo utility given in~\eqref{eq:3}, \textcolor{black}{which will be similar to a multi-horizons RL~\cite{fedus2019hyperbolic}.}

Note that the trajectories $\tau^{\theta^+}_{s_0}$ and $\tau^{\theta}$ are sampled from the learning agent's different policies. Specifically, $\tau^{\theta^+}_{s_0}$ is sampled from the most updated policy while $\tau^{\theta}$ can be sampled from the past policies during the policy optimization process. The benefit of sampling the trajectories from different polices in (\ref{eq:5}) is that we can integrate the reward function learning process with the policy search process. In particular, we can integrate the policy gradient method presented in \cite{sutton2018reinforcement} with the optimization problem in (\ref{eq:5}) and then propose a new loss function to solve the RL-MDP$\backslash\mathcal{R}$ problem as
\begin{equation}\label{eq:loss}
   \textcolor{black}{\mathcal{L(\theta, \phi)} = - (1-\rho)\mathbb{E}\left[G_{\phi}(\tau^{\theta}, \gamma) ; \pi_{\theta} \right] + \rho \left[U_{\phi}(\tau^{\theta^+}_{s_0}) - U_{\phi}(\tau^{\theta^*}_{s_0})\right],}
\end{equation}
where $\rho \in [0,1)$ is equivalent to the weight defined in~\eqref{eq:5} as $\lambda = \frac{\rho}{1-\rho}$. \textcolor{black}{Note that $\rho$ can be dynamic.} 

\subsection{Algorithm}
Based on the loss function (\ref{eq:loss}), we next will present an algorithm that concurrently updates the reward function parameter $\phi$ and the action policy parameter $\theta$. In particular, we propose a {reward and policy concurrent learning} (RPCL) algorithm that encloses the reward function learning process in the loop of policy optimization by updating $\phi$ \textcolor{black}{occasionally} with respective to $\theta$. 


\textcolor{black}{Following the philosophy that a novice needs more guidance and an expert requires less instructions, we update $\phi$ more frequently during the early learning stage and gradually reduce the frequency. \textcolor{black}{In the proposed algorithm}, we select the Fibonacci sequence as the tool to change the $\phi$ updating frequency by leveraging the growing gap between two adjacent elements in the sequence. In other words, the time gap between two updating steps in $\phi$ will increase as learning goes on, which fulfills the requirement of gradually reducing $\phi$ updating frequency. Note that other possible scaling functions can be used as long as a similar property holds. Let the $i$th updated $\phi$ be denoted as $\phi_{i}$ ($\phi_0$ means the initial $\phi$). $\phi$ remains unchanged as $\phi_{i}$ until the $(i+1)$th updation. Hence, $\phi$ is independent of $\theta$. Note that $\theta^+$ is the most updated version of $\theta$, which is also independent of the $\theta$ learning process.} Therefore, the partial partial derivative of $\mathcal{L(\theta, \phi)}$ with respect to $\theta$ is only related to the first term in (\ref{eq:loss}), \textit{i.e.}, 
$
    \frac{\partial \mathcal{L(\theta, \phi)}}{\partial \theta} = -\frac{\partial \mathbb{E}\left[G_{\phi}(\tau^{\theta}, \gamma); \pi_{\theta} \right]}{\partial \theta}.
$
\textcolor{black}{The weight $1-\rho$ is ignored as the second weighted term is not included here and the learning rate itself can incorporate the extra weight parameter.} Moreover, the gradient formulation will be the same as gradient ascent based on the discounted cumulative reward~\cite{sutton2000policy}. Therefore, we can choose an {advantage actor-critic}~\cite{mnih2016asynchronous} method to calculate $\frac{\partial \mathcal{L(\theta, \phi)}}{\partial \theta}$. Specifically, the gradient can be calculated as
$\frac{\partial \mathcal{L(\theta, \phi)}}{\partial \theta} = - \sum_{t=0}^{T-1} \triangledown_\theta \text{log} \pi_\theta (a_t|s_t) \hat{A}_t,$
where $\hat{A}_t$ denotes the advantage function and is calculated by
$\hat{A}_t = \sum_{j=t}^{T} \gamma^{j-t}g(s_j|\phi) - \hat{V}(s_t),$
and $\hat{V}(s_t)$ is the estimated value function~\cite{sutton2018reinforcement} generated by the {critic} network. After $\theta$ has been updated for \textcolor{black}{a few} times (which is determined by the $\phi$ updating frequency), the current $\theta$ is assigned to $\theta^+$ and then followed by the $\phi$ update process. Before explaining the detailed technique for updating $\phi$, it is worth mentioning the benefit of the concurrent learning structure. As the reward function is updated less frequently than the policy parameter, the percentage of required demonstrations is small. In particular, let the maximum learning episode of policy searching process be denoted as $E$. The number of demonstrations requires by RPCL is (at most) $K$ times \textcolor{black}{the total element number in the Fibonacci sequence whose value is less than $E$, \textcolor{black}{where $K$ is the number of loop times for one $\phi$ update process}}.


During the above policy learning process, the reward function is considered constant. \textcolor{black}{When the condition for $\phi$ update frequency satisfies,} we update $\phi$ for $K$ times based on the partial derivative of $\mathcal{L(\theta, \phi)}$ with respect to $\phi$ given by
\begin{equation}\label{eq:phi}
    \frac{\partial \mathcal{L(\theta, \phi)}}{\partial \phi} = -(1-\rho) \frac{\partial \mathbb{E}\left[G_{\phi}(\tau^{\theta}, \gamma); \pi_{\theta} \right]}{\partial \phi} + \rho\frac{\partial D}{\partial \phi},
\end{equation}
where \textcolor{black}{$D = U_{\phi}(\tau^{\theta^+}_{s_0}) - U_{\phi}(\tau^{\theta^*}_{s_0})$.} The expectation of the discounted cumulative reward can be approximated via samples as
$
    \mathbb{E}\left[G_{\phi}(\tau^{\theta}, \gamma); \pi_{\theta} \right] = \sum_{\tau^{\theta}} p(\tau^{\theta}) G_{\phi}(\tau^{\theta}, \gamma),
$
where $p(\tau^{\theta}) = p(\tau^{\theta}|\theta) p(\theta)$ denotes the possibility of obtaining the trajectory of $\tau^{\theta}$. Since the gradient $\frac{\partial \mathcal{L(\theta, \phi)}}{\partial \theta} $ uses an online learning technique, the probability of having $\tau^{\theta}$ for a given $\theta$ is one, \textit{i.e.}, $p(\tau^{\theta}|\theta) = 1$, and $p(\theta)$ can be simplified to be uniform. Let the number of sampled trajectories from the policy optimization process be $n$. Then we have 
$\sum_{\tau^{\theta}} p(\tau^{\theta}) G_{\phi}(\tau^{\theta}, \gamma)\approx\frac{1}{n} \sum_{k=1}^{n}G_{\phi}(\tau^{\theta_k}, \gamma),$ where $\tau^{\theta_k}$ represents the $k$th sampled trajectory from a policy $\pi_{{\theta_k}}$. Hence, the gradient of $\phi$ in (\ref{eq:phi}) can be calculated as
\begin{equation}\label{eq:reward_gradient}
    \frac{\partial \mathcal{L(\theta, \phi)}}{\partial \phi} = - \frac{1-\rho}{n} \sum_{k=1}^{n} \sum_{j=1}^{T} \gamma^{j-1}\triangledown_\phi g(s^{k}_j|\phi) + \rho \triangledown_\phi D,
\end{equation}
where $s^k_j$ is the state $s_j$ in the $k$th sampled trajectory and $n$ is the number of total sampled trajectories. The second term's derivative is calculated as 
$
\triangledown_\phi D = \frac{1}{|\Gamma|}\sum_{\gamma \in \Gamma} \big(\sum_{j=1}^{T} \gamma^{j-1}\triangledown_\phi g(s^{+}_j|\phi) - \sum_{j=1}^{T} \gamma^{j-1}\triangledown_\phi g(s^{*}_j|\phi)\big)
$
for $s^{+}_j \in \tau^{\theta^+}_{s_0}$ and $s^{*}_j \in \tau^{\theta^*}_{s_0}$. 



\section{Algorithm Evaluation} \label{sec:4}
In this section, we will demonstrate the performance of the proposed RPCL algorithm in the OpenAI gym environment~\cite{brockman2016openai} \textcolor{black}{as well as an indoor drone flight test scenario}, respectively. In particular, we test the proposed RPCL algorithm on three \textcolor{black}{gym environments}: (1) inverted pendulum with discrete control input, (2) mountain car problem~\cite{moore1990efficient} with discrete/continuous control inputs, and \textcolor{black}{(3) bipedalwalker with 4 dimensional continuous control input}. \textcolor{black}{The indoor drone flight test is conducted on our Unmanned Systems Lab's drone testing environment}. 


\subsection{Inverted Pendulum Example}
An inverted pendulum, also called cart pole, is often implemented with its pivot point mounted on a cart. The pendulum/pole can be stabilized by appropriately controlling the horizontal movement of the cart. In the OpenAI gym ({CartPole-v0}) environment, the action of the cart $a$ is discretized into two simple control inputs, \textit{i.e.}, push cart to the left ($a = 0$) and push cart to the right ($a =1$). The performance of a policy (the sequence of actions) is proportional to the episode length, \textit{i.e.}, the longer run time is, the better control policy is.  

In our experiments, we slightly modify one of the episode termination conditions with respect to the episode length so that the performance of the control policy can be verified for a longer period of time. In particular, we extend the episode length from 200 to 1000, \textit{i.e.}, one episode terminates when its length is greater than 1000. We also discard the reward value from the environment as it is assumed to be unknown in the RL-MDP$\backslash\mathcal{R}$ setting. To obtain an expert policy, we adopt the LQR method on the linearized model of the inverted pendulum and calculate the LQR gain $K$ using the parameter settings of the {CartPole-v0} environment, such that $K = [-0.9299, -2.0221, 32.3251, 11.0069].$ The control input can then be determined by the following equation,
$a_i = Ks_i, \quad i = 0, 1, \cdots, T-1,$
where $T$ is the length of that episode, $a_i$ is the $i$th step of action, and $s_i \in \mathbb{R}^{4\times 1}$ is the $i$th state that is a column vector consisted of the cart position, cart velocity, pole angle, and pole velocity at tip. To fit the control input with the {CartPole-v0} environment, we discretize $a_i$ as $a_i>0$ if $a_i \geq 0$ and $a_i=0$ otherwise. We then apply the discretized value as the action input. 


\begin{figure}[!ht]
  \centering
\begin{subfigure}[b]{0.4\linewidth}
    \includegraphics[width=\linewidth]{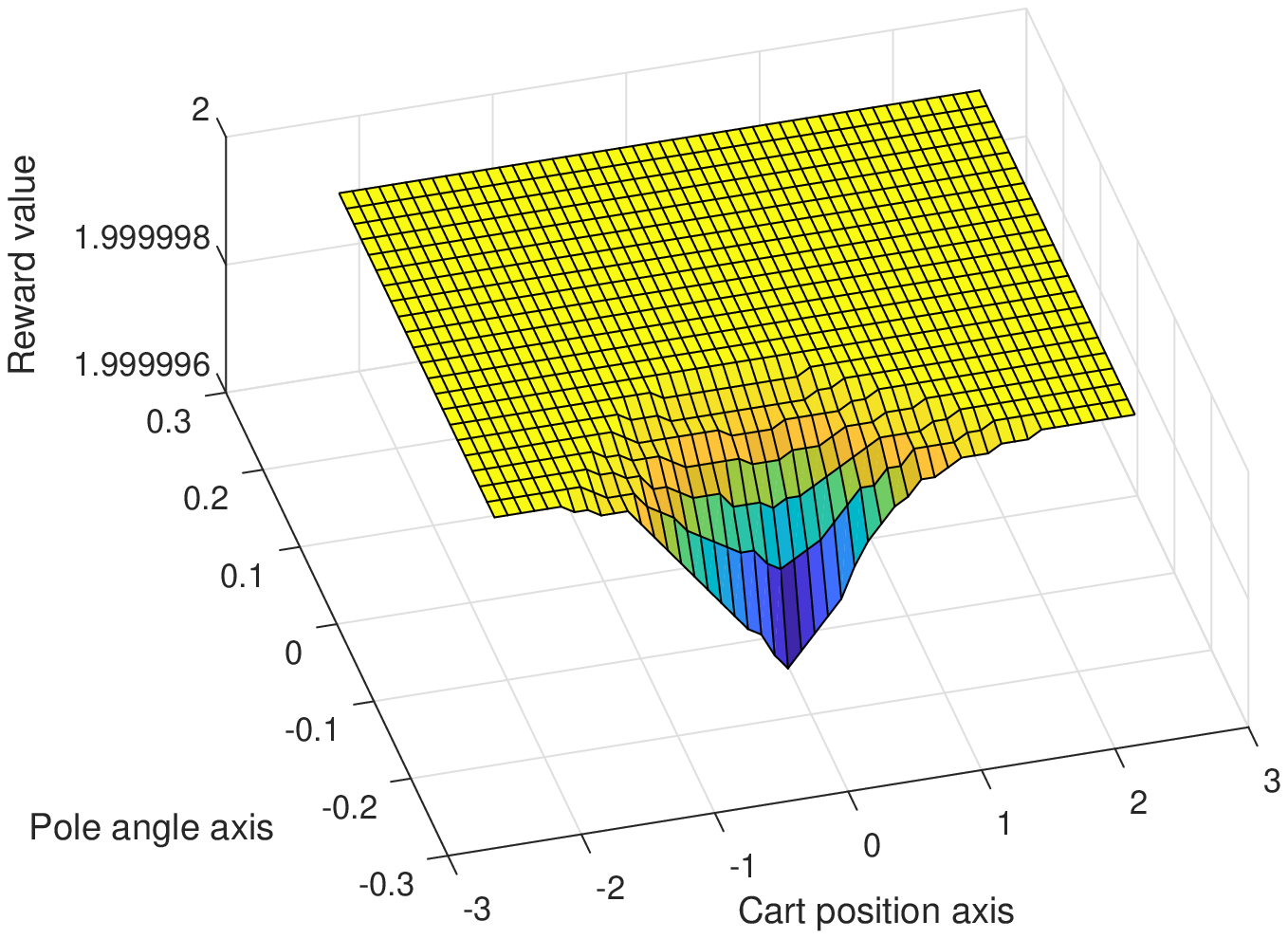}
    \caption{CartPole}
  \end{subfigure}
  \begin{subfigure}[b]{0.4\linewidth}
    \includegraphics[width=\linewidth]{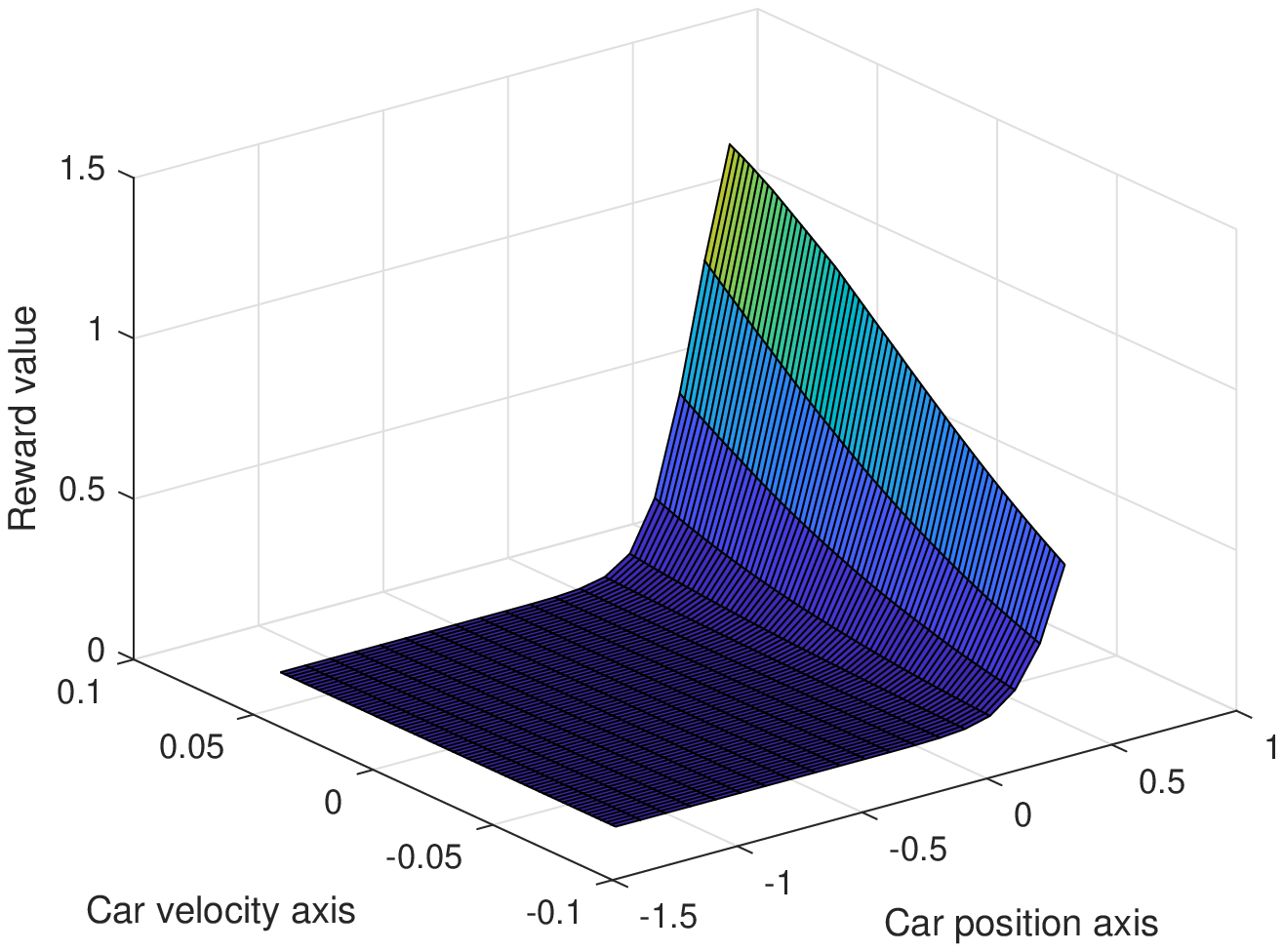}
    \caption{Mountain Car}
  \end{subfigure}
 \caption{Approximated reward function by the RPCL algorithm in the {CartPole} and Mountain Car environment.}
  \label{fig:IP_reward}
\end{figure}


\begin{figure}[!ht]
  \centering
\begin{subfigure}[b]{0.4\linewidth}
    \includegraphics[width=\linewidth]{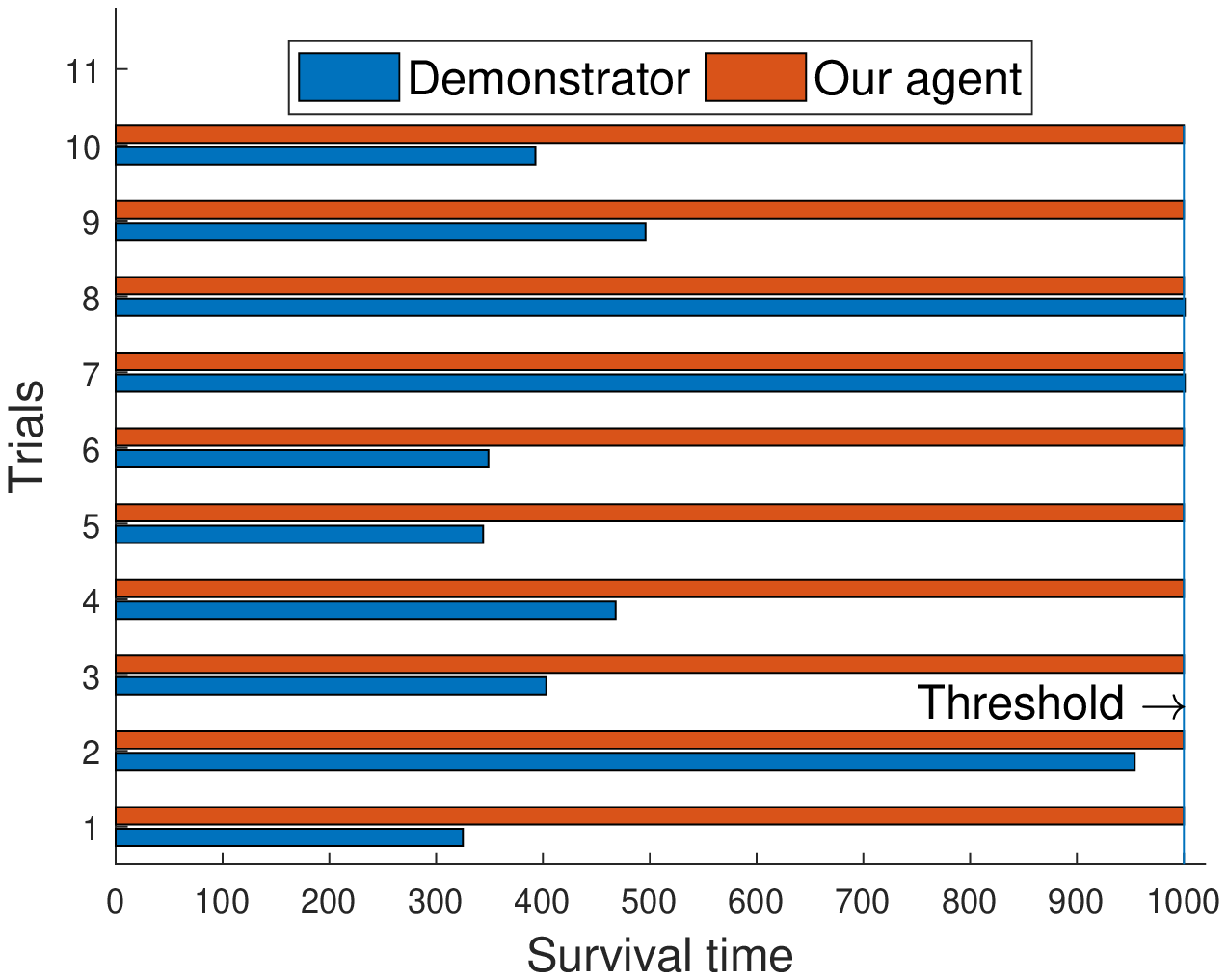}
    \caption{CartPole}
  \end{subfigure}
  \begin{subfigure}[b]{0.4\linewidth}
    \includegraphics[width=\linewidth]{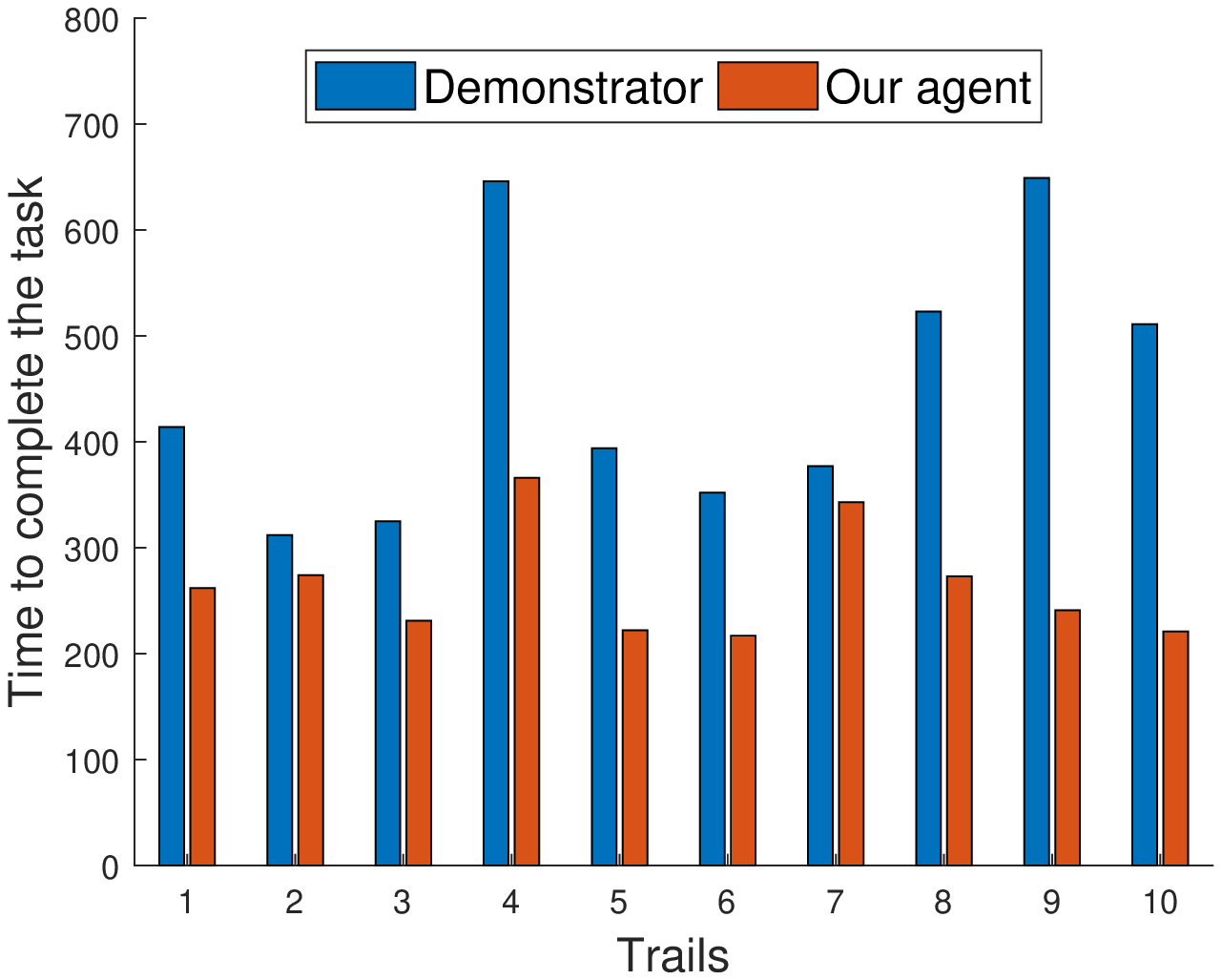}
    \caption{Mountain Car}
  \end{subfigure}
 \caption{Performance comparison between the demonstrator and the learned policy from the RPCL algorithm in the {CartPole} (discrete action) and {Mountain Car} (continuous action) environments. The demonstrator and our agent share the same initial state for each trial.}
  \label{fig:IP_performance}
\end{figure}


\textcolor{black}{With this discretized LQR demonstrator, we then implement the propose RPCL algorithm. The parameters are selected as $\rho = 0.99,~\eta = 0.99, ~\gamma = 0.995,~\Gamma = [0.9, 0.995],~e = 200,~K = 1, ~E = 5000,~\epsilon_1 = 0.1,~\epsilon_2 = 0.01,~n = 1$, and $N = 1000$.} The policy and reward functions are approximated by neural networks. In particular, the policy model consists of three layers (4-32-2). The reward function model consists of three layers (4-32-1). We select ReLU as the activation function for both models. 

\textcolor{black}{The approximated reward function derived by the RPCL algorithm is shown in Fig.\ref{fig:IP_reward}(a), where we fix the \textit{cart velocity} and \textit{pole velocity at tip} and only plot the other two dimensional variables, \textit{i.e.}, \textit{cart position} and \textit{pole angle}, with respective to the reward value. The 3D reward shape can be considered a plane, which is close to the environment reward setting (1 for every step).}

\textcolor{black}{To verify the performance of the proposed RPCL algorithm, we evaluate the learned policy against the demonstrator's policy and the policy (using the sample actor-critic RL structure as ours) learned from environment reward. In particular, we let all the three polices share the same initial state and run experiments for 1000 random initial states. From the simulation data, our RPCL agent outperforms the other two policies. The detailed statistics are provided in Table I. A random selected 10 trials of performances for both our RPCL agent and the LQR controller are shown in Fig.~\ref{fig:IP_performance}(a). For all the 10 experiments, our agent's survival time is longer than the demonstrator's.}



\subsection{Mountain Car Example}
In the mountain car example, an under-powered car learns to leverage potential energy by driving up the opposite hill before it can reach the top of the rightmost hill. In the OpenAI gym environment, there are two environments, \textit{i.e.}, the {MountainCar-v0} environment and the {MountainCarContinuous-v0} environment, which supports discrete and continuous actions respectively. In particular, {MountainCar-v0} only has three discrete action inputs, namely, push left ($a = 0$), no push ($a=1$), and push right ($a=2$), while {MountainCarContinuous-v0} allows the input action to be a continuous value. For both environments, the episode termination condition is selected as the case when the cart reaches 0.5 position (top of the rightmost hill). If the condition is not satisfied, the episode will keep on running unless other extra stop conditions are met (\textit{e.g.}, an upper limit of the experiment time). Different from the {cart-pole} problem, which considers a policy's performance proportional to the run time, the {Mountain Car} problem considers performance inverse proportional to the run time. In other words, the less steps it takes for the car to reach the 0.5 position, the better the policy is.  

We adopt the policies learned from the policy search with the original reward for both environments as the expert demonstrators since there are no analytical solutions. \textcolor{black}{In our RPCL algorithms, we set $\rho = 0.99, ~\eta = 0.99, ~\gamma =0.995, \Gamma = [0.9, 0.995],~e = 200,~K = 1, ~E = 5000,~\epsilon_1 = 0.1,~\epsilon_2 = 0.01,~n = 1$, and $N = 1000$. The policy model is approximated by a 3-layer neural network (2-128-2 for continuous action and 2-128-1 for discrete action). The reward function is approximated by a 3-layer neural network (2-32-1). We select ReLU as the activation function for both networks.}

\textcolor{black}{The approximated reward function from our RPCL method is shown in Fig.\ref{fig:IP_reward}(b). As the mountain car wants to accomplish the task as soon as possible, the reward value for major states is zero. There are non-zero values only when the car position is close to 0.5, which is the termination condition. In addition, we can observe that the larger the car velocity is, the higher the reward value is. This is sound since a higher velocity means a higher kinetic energy and a higher possibility to drive up.}

\textcolor{black}{Similar to the tests in the inverted pendulum, we evaluate the learned policy against the demonstrator's policy and the policy (using the sample actor-critic RL structure as ours) learned from environment reward. In particular, we let all three polices share the same initial state and run experiments for 1000 random initial states. According to the simulation data, our RPCL agent provides better results than the other two policies. A randomly selected 10 trials of performances for both our RPCL agent and the demonstrator are shown in Fig.~\ref{fig:IP_performance}(b), which shows that our agent takes less time to complete the task than the demonstrator. The detailed statistics of the evaluation are provided in Table 1.} \textcolor{black}{We also explore the choice of different discount factor sets and the corresponding effect on the agent's performance. From the results shown in Table 2, we can see that the inclusion of more discount factors in the form of the proposed stereo utility can generally yield improved performance. In particular, for the CartPole environment, the selection of discount factor sets does not impact the performance. This is because CartPole has a short-term goal (i.e., the pendulum does not fall down). For both MountainCar-discrete and MountainCar-continuous, the selection of a diverse discount factor set is beneficial. Hence, the adoption of the proposed stereo utility is one critical factor for agents to learn and then exceed demonstrators, especially for tasks with long-term goals.}

\begin{minipage}{1\linewidth}
\label{tab: tau_set}
\centering
\captionof{table}{Results comparison for {CartPole-v0} (CP), {MountainCar-v0} (MC), and {MountainCarContinuous-v0} (MCC) with respect to the running steps}
\begin{tabular}{c|c|c|c}
\toprule[1pt]
Environment  & RPCL            & Expert   & AC with Env Reward  \\\hline
CP             & 1000$\pm$0          &   716$\pm$310                 & 972 $\pm$148                   \\\hline
MC            & 135$\pm$22          &  253$\pm$143            & 135$\pm$32              \\\hline
MCC       & 273$\pm$61          &  417$\pm$127            & 393$\pm$120               \\\hline
\end{tabular}
\end{minipage}
\newline

\begin{minipage}{1\linewidth}
\label{tab: tau_set}
\centering
\captionof{table}{Effect of the choice of discount factor set $\Gamma$}
\begin{tabular}{c|c|c|c}
\toprule[1pt]
\textbf{$\Gamma$}  & CartPole            & MountainCar-discrete   & MountainCar-continuous  \\\hline
[0.9]              & 1000$\pm$0          &   Fail                 & 384$\pm$103                   \\\hline
[0.995]            & 998$\pm$38          &  146$\pm$44            & 278$\pm$71              \\\hline
[0.9, 0.995]       & 1000$\pm$0          &  135$\pm$22            & 273$\pm$61               \\\hline
[0.9, 0.99, 0.995] & 1000$\pm$0          &  135$\pm$32            & 261$\pm$70               \\\hline
\end{tabular}
\end{minipage}



\subsection{BipedalWalker Example}

\begin{figure}[!ht]
  \centering
  \begin{subfigure}[b]{0.19\linewidth}
    \includegraphics[width=\linewidth]{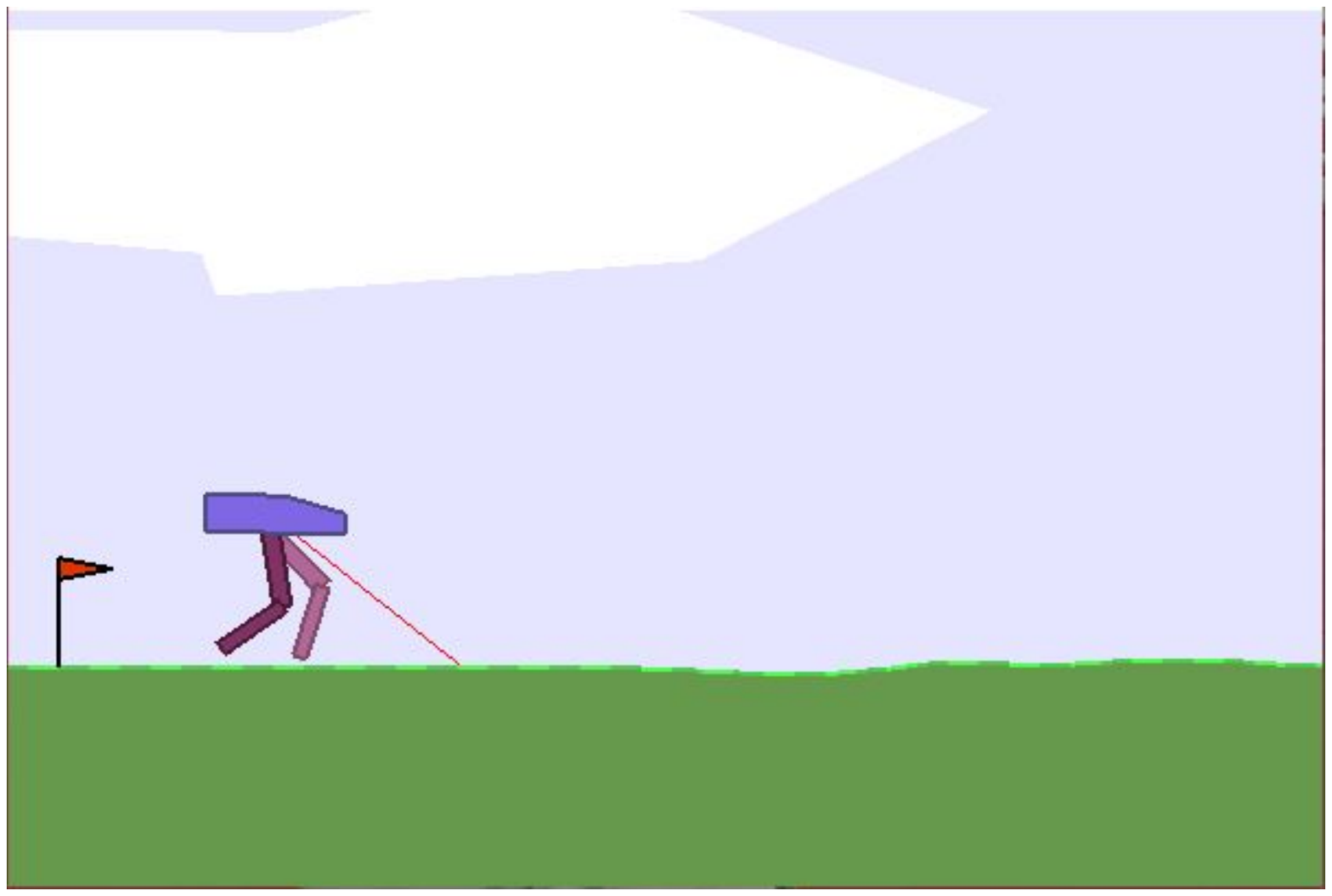}
  \end{subfigure}
  \begin{subfigure}[b]{0.19\linewidth}
    \includegraphics[width=\linewidth]{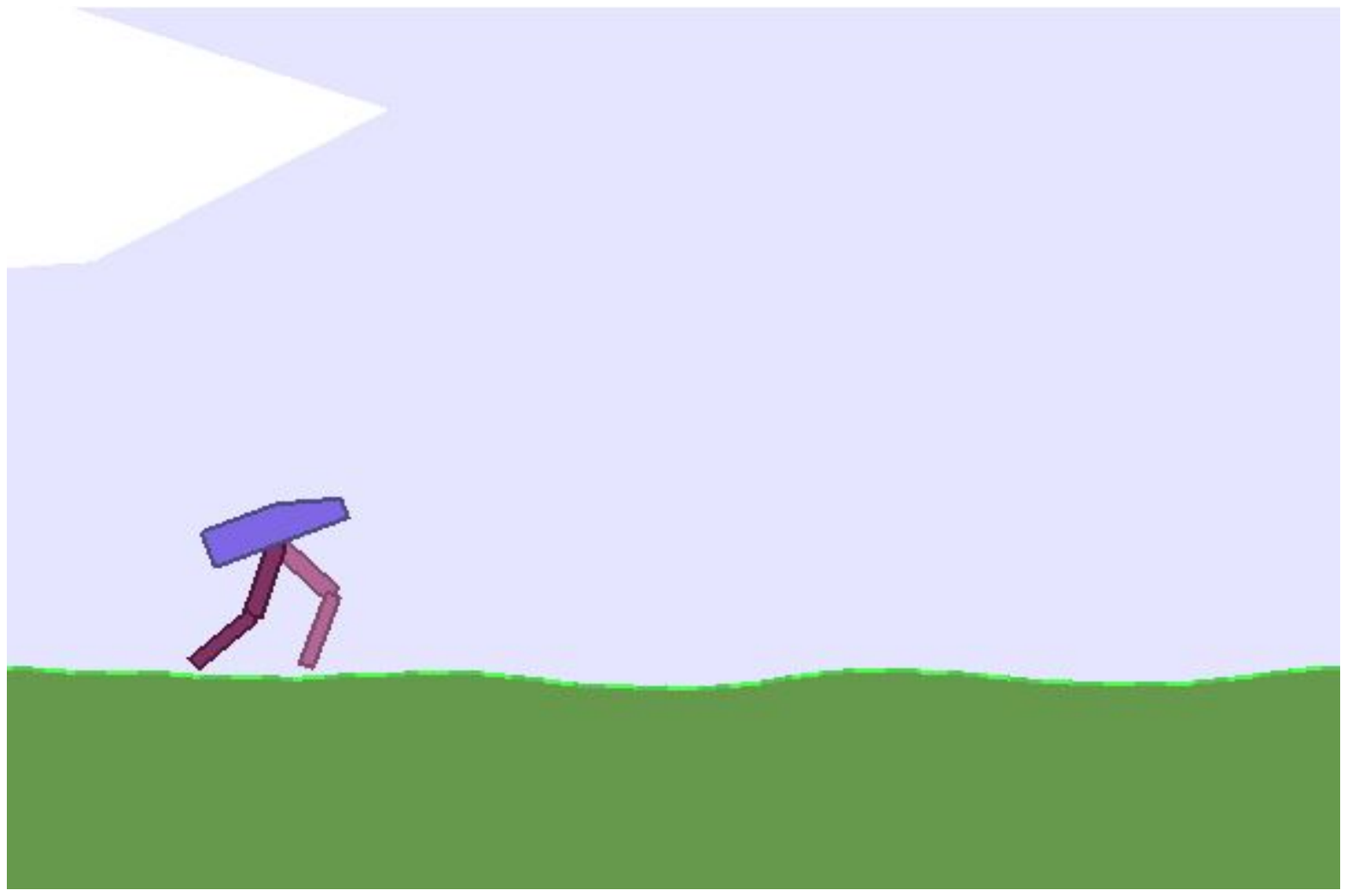}
  \end{subfigure}
  \begin{subfigure}[b]{0.19\linewidth}
    \includegraphics[width=\linewidth]{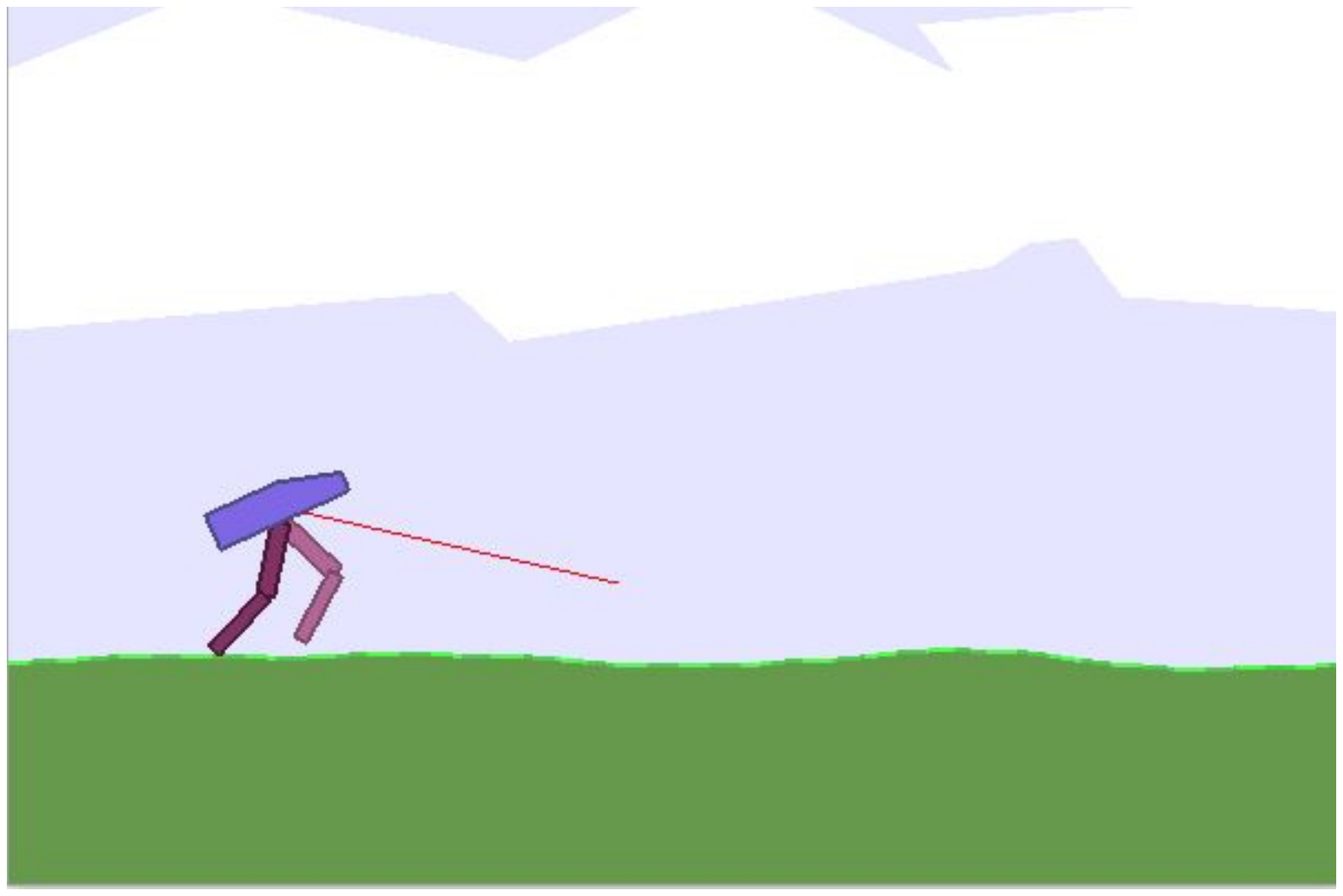}
  \end{subfigure}
  \begin{subfigure}[b]{0.19\linewidth}
    \includegraphics[width=\linewidth]{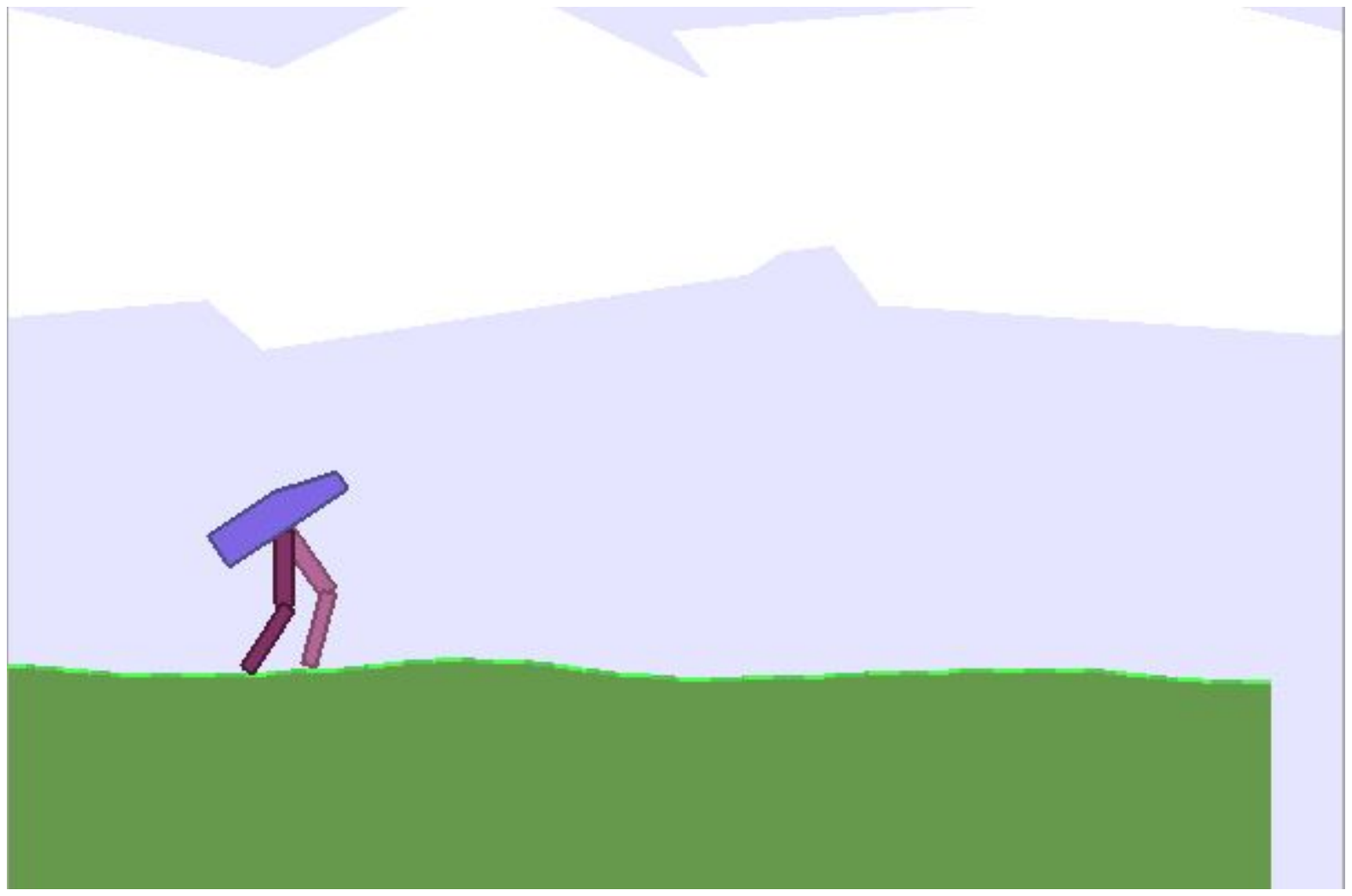}
  \end{subfigure}
  \begin{subfigure}[b]{0.19\linewidth}
    \includegraphics[width=\linewidth]{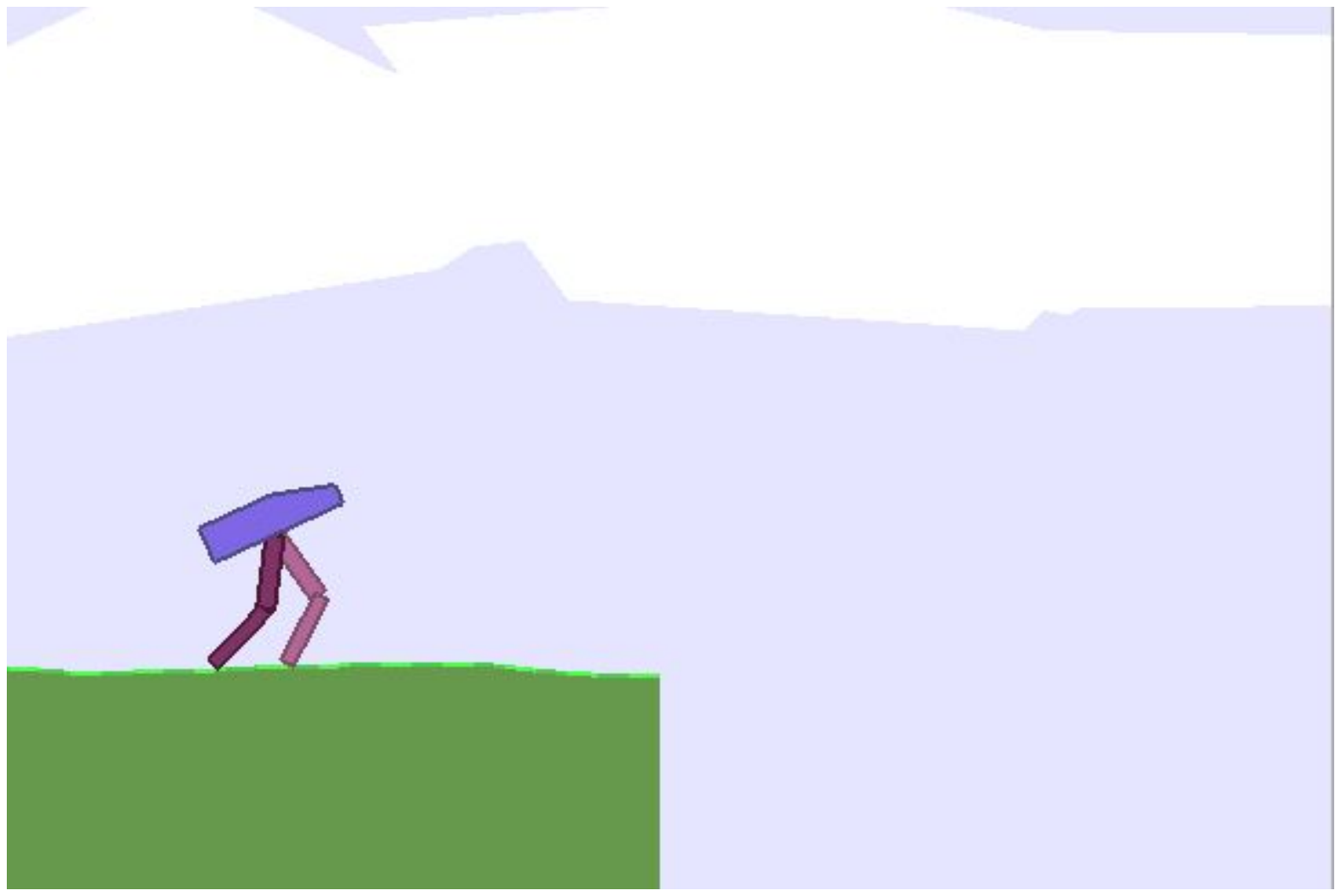}
  \end{subfigure}
  \caption{One trajectory generated via the learned policy from the proposed RPCL method.}
  \label{fig:bipedalwalker}
\end{figure}

\textcolor{black}{Comparing to the aforementioned two low-dimensional gym environments, the {BipedalWalker-v3} environment has a higher dimensional task to be solved. In particular, the {BipedalWalker-v3} environment simulates the bipedal locomotion, which takes 4 degrees of freedom (2 hip and knee joints) action inputs and provides 24-dimensional state information. The information consists of hull angle speed, angular velocities, horizontal speed, vertical speed, position of joints and joints angular speed, legs contact with ground, and $10$ lidar rangefinder measurements. The performance of the agent is not directly related to the run time. Instead, the performance is measured by whether the agent moves to the far end without falling and how much motor torque needs to be applied.} 

\textcolor{black}{In our RPCL simulation, we adopt a policy learned from the policy search with the original reward, obtained via DDPG method \cite{lillicrap2015continuous}, as the expert demonstrators. The demonstrator's performance is $288\pm69$ evaluated on 500 rollouts. {For our RPCL algorithm, we set $\rho = 0.99, ~\eta = 0.99, ~\gamma =0.99, \Gamma = [0.9, 0.99, 0.995],~e = 200,~K = 200, ~E = 2000,~\epsilon_1 = 0.001,~\epsilon_2 = 0.0003,~n = 1$, and $N = 1000$. The policy model is approximated by a 3-layer neural network (24-256-8). The reward function is approximated by a 4-layer neural network (28-64-32-1). We select ReLU as the activation function for both networks.}}

\textcolor{black}{A trajectory of the agent generated by the learned policy from our RPCL algorithm is shown in Fig. \ref{fig:bipedalwalker}, which demonstrates the successful locomotion of the agent. For a 500-rollout statistical evaluation, our RPCL policy can get a score of 298$\pm$27, which outperforms the demonstrator.}

To further show the advantages of the proposed RPCL method, we also perform comparison of the proposed RPCL method with other baseline methods, including behavior cloning, generative adversarial imitation learning (GAIL), maximum margin inverse reinforcement learning, and maximum-entropy inverse reinforcement learning. Table 2 shows the outcomes of the conducted comparisons. It can be seen that our RPCL algorithm outperforms Demonstrator, Behavior cloning, Maximum Margin, RL with environment reward, and GAIL in all examples. We were only able to obtain one example (MountainCar-discrete) that maximum-entropy IRL applies. The maximum-entropy IRL works better than ours, \textcolor{black}{while requiring 5 times more episodes of training.}

\begin{minipage}{1\linewidth}
\label{tab: compare}
\centering
\captionof{table}{Comparison with other baseline methods using the cumulative environment reward}
\begin{threeparttable}
\begin{tabular}{c|c|c|c|c}
\toprule[1pt]
\textbf{IRL methods}        & CartPole            & MountainCar-discrete   & MountainCar-continuous  &  BipedalWalker    \\\hline
Behaviour Cloning          & 886$\pm$183         &  Fails                 &   72$\pm$41        &   287$\pm$75         \\\hline
RPCL                        & {1000$\pm$0} &  -143$\pm$36           & {84$\pm$6}     &  {298$\pm$27}  \\\hline
RL with Env reward          & 972$\pm$148         &  -162$\pm$41           &   82$\pm$13      &   {290$\pm$25}     \\\hline
GAIL                        & 36$\pm$22$^*$       &    N/A                 &  N/A        &  255$\pm$123         \\\hline
Maximum Margin              & 983$\pm$122         &  -153$\pm$34            &  81$\pm$15    &  274$\pm$1                      \\\hline
MaxEnt                      &  N/A                & {-123$\pm$ 1}$^1$   &  N/A       &    N/A               \\\bottomrule[1.25pt]
\textbf{Demonstrator}       & 798$\pm$287         & -247$\pm$135           & 76$\pm$8       &    288$\pm$69             \\\hline
\end{tabular}


\begin{tablenotes}
\small
\item [*] obtained in the wrapped environment setting (maximum 200 steps).
\item [1] requires 30000 episodes of training.
\end{tablenotes}
\end{threeparttable}
\end{minipage}

\subsection{Indoor Real Drone Testing Example}
\textcolor{black}{To verify the performance of our RPCL algorithm in real-world experiments, we test it on an indoor drone testing environment, where the environment rewards are unavailable. In particular, we conduct a simple task by driving the drone from a given ground position (-1.2$\pm$0.1m, 0.2$\pm$0.1m, 0) to the origin of 3D coordinate system, \textit{i.e.}, (0, 0, 0). The control input for the drone is discretized into seven possible actions, \textit{i.e.}, stay still, move forward/backward, move left/right, and rise/fall.}

\textcolor{black}{We choose a PI controller as the demonstrator. More precisely, the PI controller will first navigate the drone parallel to the ground when the drone takes off. The movement in z axis is only activated when the drone is above the acceptable landing area. In our tests, we set the acceptable landing area as a circle with a radius of 0.25 meters around the origin. To train our RPCL algorithm, we select an actor-critic algorithm and take the $x,y,z$ positions and $v_x, v_y, v_z$ velocities as the input states. The parameters of our RPCL are set as $\rho = 0.99, ~\eta = 0.99, ~\gamma =0.99, \Gamma = [0.9, 0.995],~\epsilon_1 = 5e-5,~\epsilon_2 = 0.01,~n = 1$, and $N = 1000$. The variables $e$ and $E$ are not set because operating large real flight tests is time-consuming. Hence, the stop condition is determined by the operator. In our experiments, we conduct 279 episodes of flight and only 14 demonstrations are required.}

\begin{figure}[!ht]
  \centering
  \begin{subfigure}[b]{0.3\linewidth}
    \includegraphics[width=\linewidth]{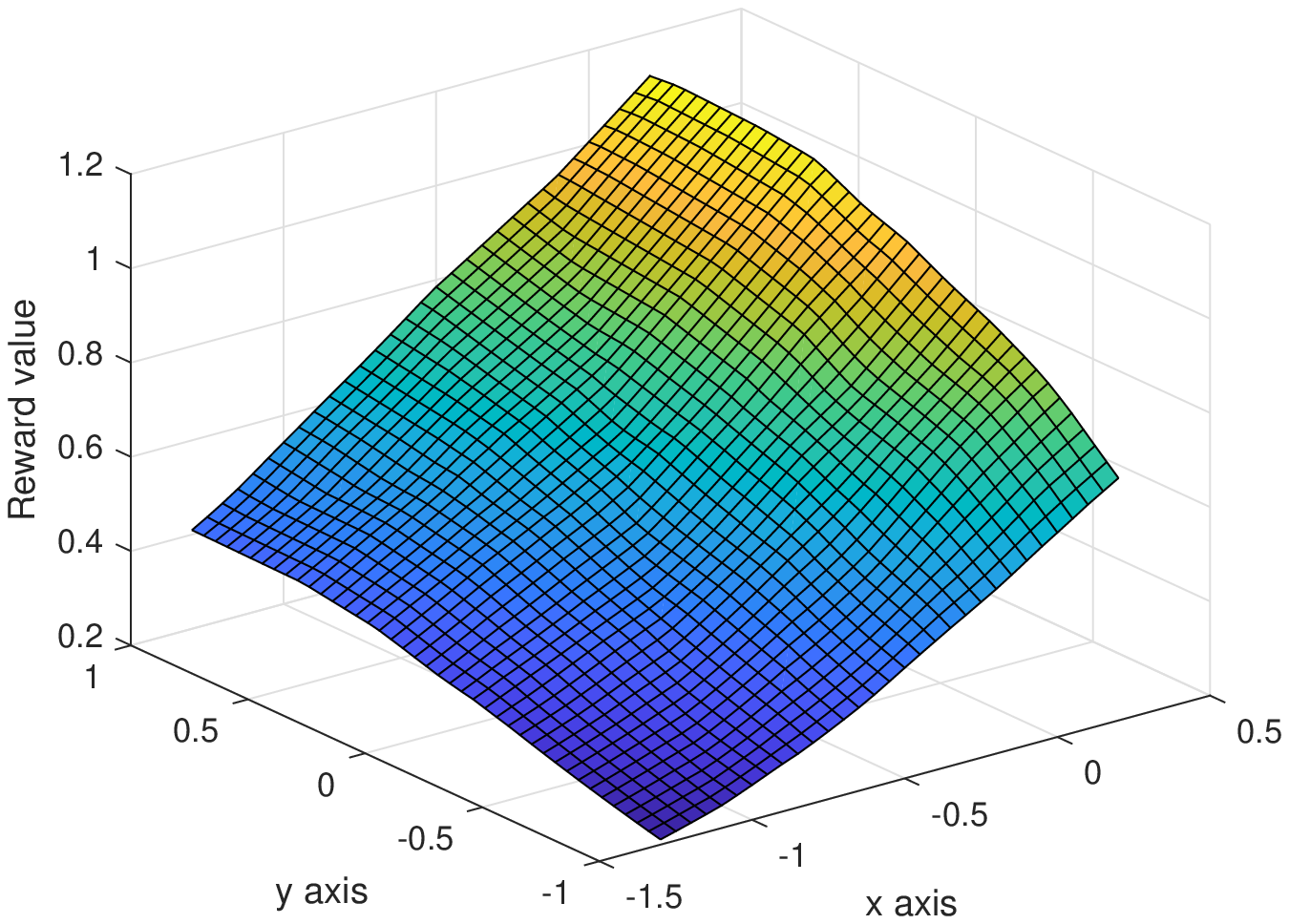}
     \caption{Height = 0.01m}
  \end{subfigure}
  \begin{subfigure}[b]{0.3\linewidth}
    \includegraphics[width=\linewidth]{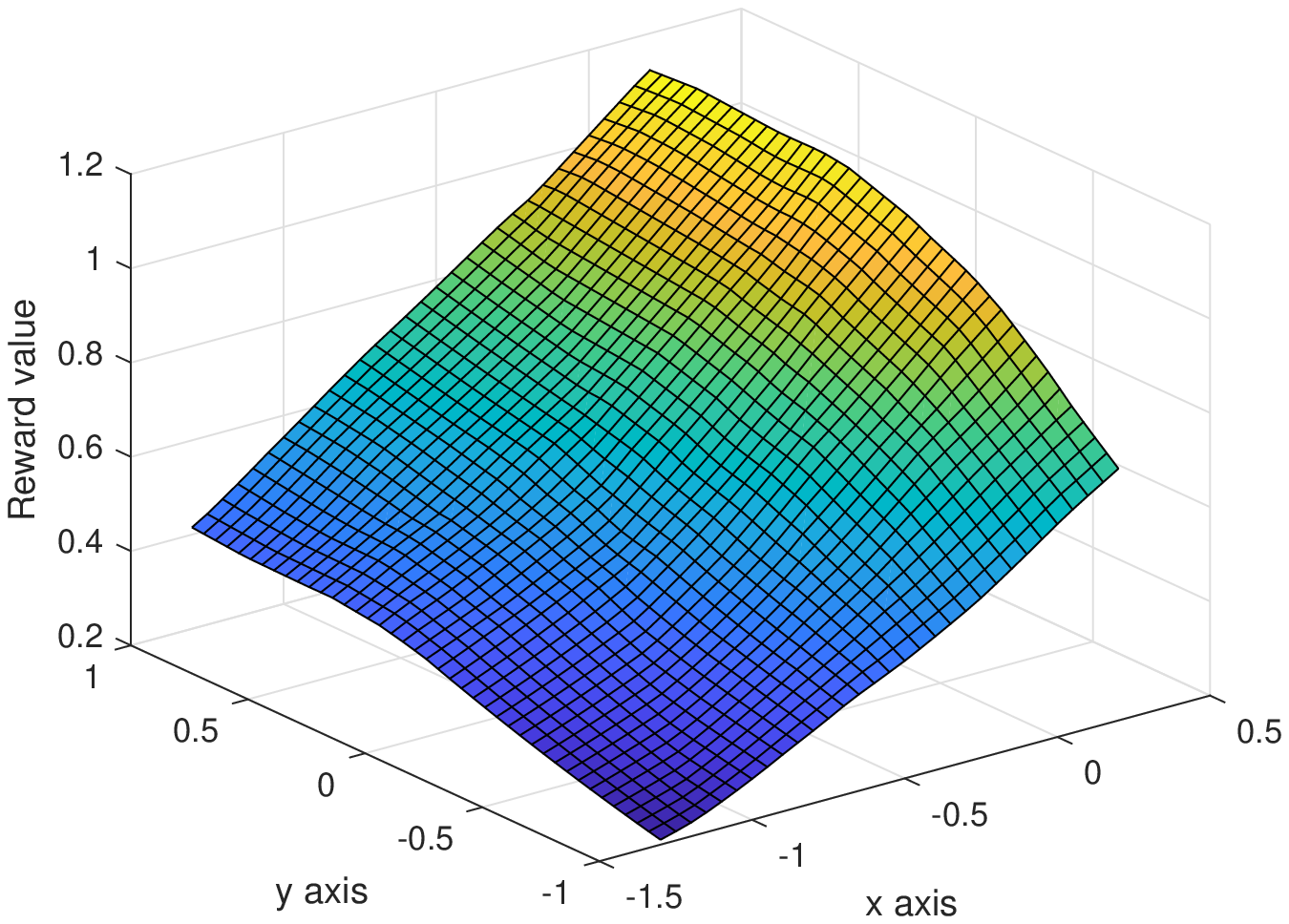}
    \caption{Height = 0.2m}
  \end{subfigure}
  \begin{subfigure}[b]{0.3\linewidth}
    \includegraphics[width=\linewidth]{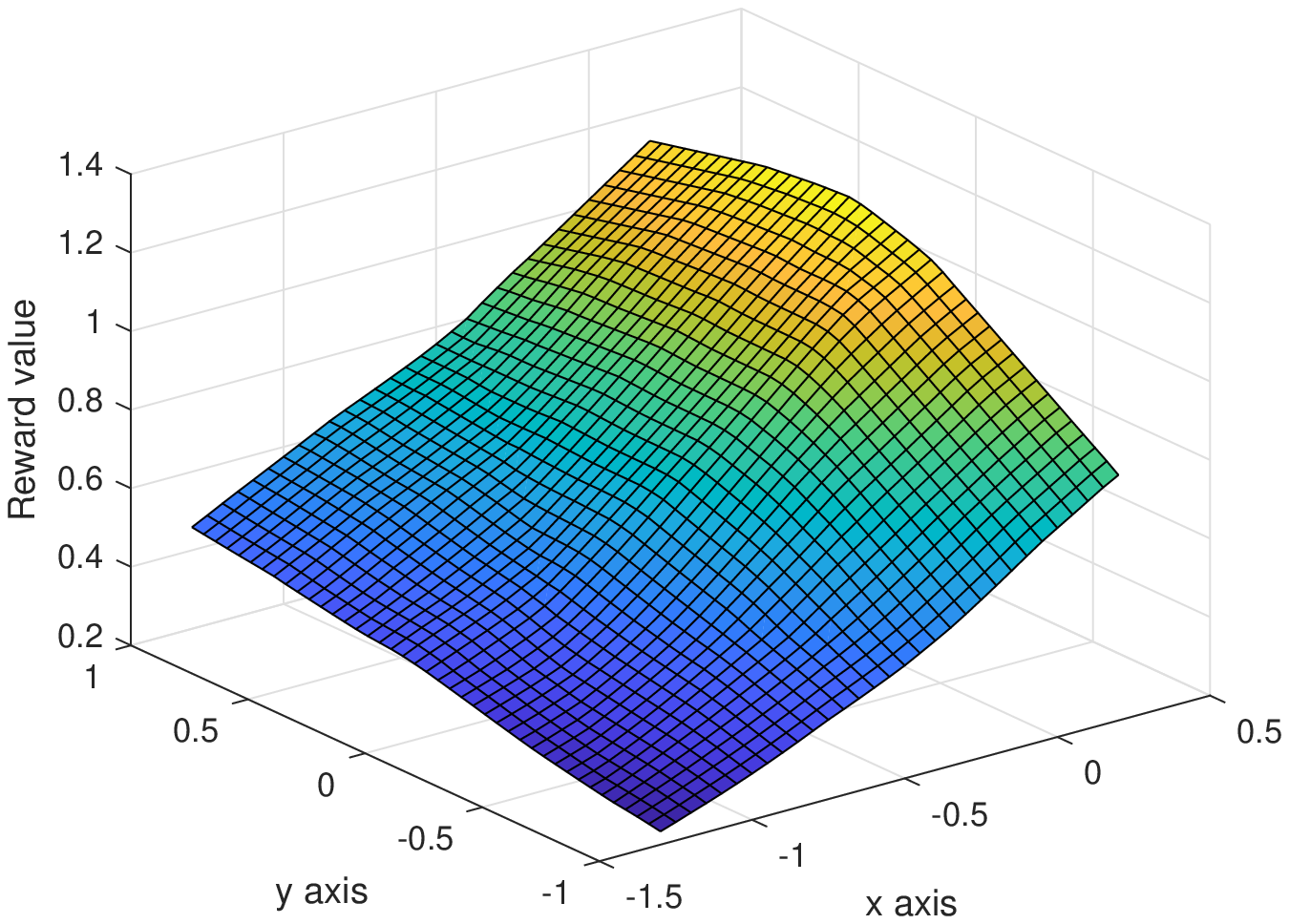}
    \caption{Height = 0.8m}
  \end{subfigure}
  \caption{Horizontal plane reward plot at different heights.}
  \label{fig:drone_reward}
\end{figure}


\begin{figure}
  \begin{center}
    \includegraphics[width=0.5\textwidth]{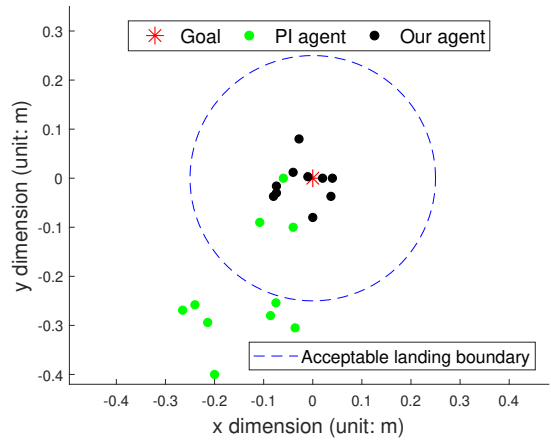}
  \end{center}
  \caption{Performance comparison between the demonstrator and the RPCL algorithm for indoor real drone testing.}
  \label{fig:MC_performance}
\end{figure}
\textcolor{black}{The approximated reward function is shown in Fig. \ref{fig:drone_reward}, where each subplot represents the horizontal plane with respect to different heights (z-axis). From those three reward plots, it can be observed that the reward is gradually increasing to the goal area. In addition, the rewards in different heights follow similar trends albeit vary at different heights.}

\textcolor{black}{When the number of the training episode number reaches 270, the drone already performs well. Hence, we stop the training at episode 279 and then compare the derived policy with the PI controller. 10 flights are conducted and the landing markers for our agent and the PI controller are plotted in Fig. \ref{fig:MC_performance}. It can be seen that our agent performs better and more stable than the PI controller does. The average landing location for our agent is $(-0.02, -0.01)$ with a standard deviation 0.04, while the average landing location for the PI controller is $(-0.13, -0.23)$ with a standard deviation 0.18.}


\section{Conclusion} \label{sec:5}
In this paper, we proposed a new reward and policy concurrent learning (RPCL) algorithm to derive a control policy that can mimic and outperform expert's demonstrations in Markov decision processes, where the reward function is unknown. The RPCL algorithm is built based on the construction of a new stereo utility function and the design of a new loss function. We presented the proposed RPCL algorithm, including its motivation, formulation, and algorithm. We also conducted experiment studies in three OpenAI environments and an indoor drone flight environment to show the effectiveness of the proposed RPCL algorithm. 




\bibliography{ref}  

\begin{thebibliography}{10}

\bibitem{silver2016mastering}
David Silver, Aja Huang, Chris~J Maddison, Arthur Guez, Laurent Sifre, George
  Van Den~Driessche, Julian Schrittwieser, Ioannis Antonoglou, Veda
  Panneershelvam, Marc Lanctot, et~al.
\newblock Mastering the game of go with deep neural networks and tree search.
\newblock {\em Nature}, 529(7587):484, 2016.

\bibitem{mnih2015human}
Volodymyr Mnih, Koray Kavukcuoglu, David Silver, Andrei~A Rusu, Joel Veness,
  Marc~G Bellemare, Alex Graves, Martin Riedmiller, Andreas~K Fidjeland, Georg
  Ostrovski, et~al.
\newblock Human-level control through deep reinforcement learning.
\newblock {\em Nature}, 518(7540):529, 2015.

\bibitem{williams1992simple}
Ronald~J Williams.
\newblock Simple statistical gradient-following algorithms for connectionist
  reinforcement learning.
\newblock {\em Machine Learning}, 8(3-4):229--256, 1992.

\bibitem{schulman2017proximal}
John Schulman, Filip Wolski, Prafulla Dhariwal, Alec Radford, and Oleg Klimov.
\newblock Proximal policy optimization algorithms.
\newblock {\em arXiv:1707.06347}, 2017.

\bibitem{ng1999policy}
Andrew~Y Ng, Daishi Harada, and Stuart Russell.
\newblock Policy invariance under reward transformations: Theory and
  application to reward shaping.
\newblock In {\em International Conference on Machine
  Learning (ICML)}, pages 278--287, 1999.

\bibitem{kumar2018bipedal}
Arun Kumar, Navneet Paul, and SN~Omkar.
\newblock Bipedal walking robot using deep deterministic policy gradient.
\newblock {\em arXiv:1807.05924}, 2018.

\bibitem{everitt2017reinforcement}
Tom Everitt, Victoria Krakovna, Laurent Orseau, Marcus Hutter, and Shane Legg.
\newblock Reinforcement learning with a corrupted reward channel.
\newblock {\em arXiv:1705.08417}, 2017.

\bibitem{huang2019deceptive}
Yunhan Huang and Quanyan Zhu.
\newblock Deceptive reinforcement learning under adversarial manipulations on
  cost signals.
\newblock {\em arXiv:1906.10571}, 2019.

\bibitem{ng2000algorithms}
Andrew~Y Ng, Stuart~J Russell, et~al.
\newblock Algorithms for inverse reinforcement learning.
\newblock In {\em International Conference on Machine
  Learning (ICML)}, volume~1, page~2, 2000.

\bibitem{ziebart2008maximum}
Brian~D Ziebart, Andrew Maas, J~Andrew Bagnell, and Anind~K Dey.
\newblock Maximum entropy inverse reinforcement learning.
\newblock {\em AAAI Conference on Artificial Intelligence},
  2008.

\bibitem{finn2016guided}
Chelsea Finn, Sergey Levine, and Pieter Abbeel.
\newblock Guided cost learning: Deep inverse optimal control via policy
  optimization.
\newblock In {\em International Conference on Machine
  Learning (ICML)}, pages 49--58, 2016.

\bibitem{schaal1999imitation}
Stefan Schaal.
\newblock Is imitation learning the route to humanoid robots?
\newblock {\em Trends in Cognitive Sciences}, 3(6):233--242, 1999.

\bibitem{ho2016generative}
Jonathan Ho and Stefano Ermon.
\newblock Generative adversarial imitation learning.
\newblock In {\em Advances in Neural Information Processing Systems}, pages
  4565--4573, 2016.

\bibitem{syed2008game}
Umar Syed and Robert~E Schapire.
\newblock A game-theoretic approach to apprenticeship learning.
\newblock In {\em Advances in Neural Information Processing Systems}, pages
  1449--1456, 2008.

\bibitem{zhifei2012review}
Shao Zhifei and Er~Meng Joo.
\newblock A review of inverse reinforcement learning theory and recent
  advances.
\newblock In {\em 2012 IEEE Congress on Evolutionary Computation}, pages 1--8,
  2012.

\bibitem{abbeel2004apprenticeship}
Pieter Abbeel and Andrew~Y Ng.
\newblock Apprenticeship learning via inverse reinforcement learning.
\newblock In {\em International Conference on Machine
  Learning (ICML)}, page~1, 2004.

\bibitem{levine2011nonlinear}
Sergey Levine, Zoran Popovic, and Vladlen Koltun.
\newblock Nonlinear inverse reinforcement learning with gaussian processes.
\newblock In {\em Advances in Neural Information Processing Systems}, pages
  19--27, 2011.

\bibitem{wulfmeier2015maximum}
Markus Wulfmeier, Peter Ondruska, and Ingmar Posner.
\newblock Maximum entropy deep inverse reinforcement learning.
\newblock {\em arXiv:1507.04888}, 2015.

\bibitem{arora2018survey}
Saurabh Arora and Prashant Doshi.
\newblock A survey of inverse reinforcement learning: Challenges, methods and
  progress.
\newblock {\em arXiv:1806.06877}, 2018.

\bibitem{sutton2018reinforcement}
Richard~S Sutton and Andrew~G Barto.
\newblock {\em Reinforcement learning: An introduction}.
\newblock MIT press, 2018.

\bibitem{russell1998learning}
Stuart~J Russell.
\newblock Learning agents for uncertain environments.
\newblock In {\em Annual Conference on Learning Theory}, pages
  101--103, 1998.

\bibitem{shiarlis2016inverse}
Kyriacos Shiarlis, Joao Messias, and Shimon Whiteson.
\newblock Inverse reinforcement learning from failure.
\newblock In {\em Proceedings of the 2016 International Conference on
  Autonomous Agents \& Multiagent Systems}, pages 1060--1068, 2016.

\bibitem{ratliff2006maximum}
Nathan~D Ratliff, J~Andrew Bagnell, and Martin~A Zinkevich.
\newblock Maximum margin planning.
\newblock In {\em International Conference on Machine
  Learning (ICML)}, pages 729--736, 2006.

\bibitem{fedus2019hyperbolic}
William Fedus, Carles Gelada, Yoshua Bengio, Marc~G Bellemare, and Hugo
  Larochelle.
\newblock Hyperbolic discounting and learning over multiple horizons.
\newblock {\em arXiv preprint arXiv:1902.06865}, 2019.

\bibitem{sutton2000policy}
Richard~S Sutton, David~A McAllester, Satinder~P Singh, and Yishay Mansour.
\newblock Policy gradient methods for reinforcement learning with function
  approximation.
\newblock In {\em Advances in Neural Information Processing Systems}, pages
  1057--1063, 2000.

\bibitem{mnih2016asynchronous}
Volodymyr Mnih, Adria~Puigdomenech Badia, Mehdi Mirza, Alex Graves, Timothy
  Lillicrap, Tim Harley, David Silver, and Koray Kavukcuoglu.
\newblock Asynchronous methods for deep reinforcement learning.
\newblock In {\em International Conference on Machine
  Learning (ICML)}, pages 1928--1937, 2016.

\bibitem{brockman2016openai}
Greg Brockman, Vicki Cheung, Ludwig Pettersson, Jonas Schneider, John Schulman,
  Jie Tang, and Wojciech Zaremba.
\newblock OpenAI gym.
\newblock {\em arXiv:1606.01540}, 2016.

\bibitem{moore1990efficient}
Andrew~William Moore.
\newblock Efficient memory-based learning for robot control.
\newblock 1990.

\bibitem{lillicrap2015continuous}
Timothy~P Lillicrap, Jonathan~J Hunt, Alexander Pritzel, Nicolas Heess, Tom
  Erez, Yuval Tassa, David Silver, and Daan Wierstra.
\newblock Continuous control with deep reinforcement learning.
\newblock {\em arXiv preprint arXiv:1509.02971}, 2015.

\end{thebibliography}
\bibliographystyle{unsrt}

\clearpage

\section*{Appendix 1: Pseudocode for the Proposed RPCL Method}

The pseudocode for the proposed RPCL method is given below.

\begin{algorithm}
\caption{Reward and Policy Concurrent Learning}
\label{alg:1}
\begin{algorithmic}[1]
\State Initialize reward and policy parameters $\phi, \theta$
\State Initialize sample inventory $\mathcal{M}$ to capacity $N$
\State Set learning rates $\epsilon_1$ and $\epsilon_2$ for $\phi$ and $\theta$
\State \textcolor{black}{Generate the Fibonacci sequence $F = [0,1,1,2,3,5,\cdots]$}
\State Set ratio $\rho$ and its decay rate $\eta$
\State Set the sampled number $n \,(n < N)$
\State Set the discount factor $\gamma$ and $\Gamma$
\State Set minimum episode $e$, loop stop condition
\State Set the maximum learning episode $E$
\State set index = 0
\For {episode $=1,\cdots, E$} 
    \If {episode $> e$ and stop condition}  
         \State \textbf{break}      
    \EndIf
    \State Initialize state $s_0$
    \State Obtain trajectory: $\tau^{\theta} = \{s_0, \pi_{\theta}(a_0|s_0), g(s_1|\phi), s_1, ..., s_T\}$   
    
   \State Calculate the advantage function $\hat{A}_t$
   \State $\theta \leftarrow \theta + \epsilon_2 \sum_{t=0}^{T-1} \triangledown_\theta \text{log} \pi_\theta (a_t|s_t) \hat{A}_t$ 
   \State Calculate the discounted cumulative reward $G_{\phi}(\tau^{\theta}, \gamma)$
   \State Store $G_{\phi}(\tau^{\theta}, \gamma)$ into $\mathcal{M}$
   \If {episode $\geq$ $F$[index]}
        \State index $\leftarrow$ index+1
        \State $\theta^+ \leftarrow \theta$
        \Repeat
        \State Sample $n$ cumulative rewards from $\mathcal{M}:$ $\{G_{\phi}(\tau^{\theta_1}, \gamma), G_{\phi}(\tau^{\theta_2}, \gamma), ... G_{\phi}(\tau^{\theta_n}, \gamma)\}$  
        \State Initialize state $s_0$ 
        \State Obtain one trajectory $\tau^{\theta^+}_{s_0}$ under current policy $\theta^+$
        \State Obtain a demonstration $\tau^{\theta^*}_{s_0}$ 
        \State Calculate the {stereo utility} difference $D = U_{\phi}(\tau^{\theta^+}_{s_0}) - U_{\phi}(\tau^{\theta^*}_{s_0})$
        \State $\phi \leftarrow  \phi-\epsilon_1 \left(\frac{1-\rho}{n} \sum_{k=1}^{n} -\triangledown_\phi G_{\phi} (\tau^{\theta_{k}}, \gamma) + \rho \triangledown_\phi D \right)$
        \Until{$K$ times}
        \State $\rho \leftarrow \rho * \eta$
   \EndIf 
\EndFor
  
\end{algorithmic}
\label{Alg:topology_update}  
\end{algorithm}

\end{document}